%% file: acl_latex.tex
\documentclass[11pt]{article}

\usepackage[final]{acl}

\usepackage[table]{xcolor}
\usepackage{times}
\usepackage{latexsym}
\usepackage{float}
\usepackage{amsmath}
\usepackage{amssymb}
\usepackage{mathrsfs} 
\usepackage[most,skins,theorems]{tcolorbox}  
\usepackage{xcolor}      
\usepackage{multirow}
\usepackage{pifont}
\usepackage{booktabs}
\tcbuselibrary{breakable,skins}
\usepackage[T1]{fontenc}
\usepackage{enumitem}
\usepackage{afterpage}
\usepackage[utf8]{inputenc}

\usepackage{microtype}
\usepackage{makecell}
\usepackage{stfloats}
\usepackage{inconsolata}
\usepackage{caption}
\usepackage{graphicx}
\usepackage{titlesec}

\tcbset{
  aiboxstyle/.style={
    width=\linewidth,
    top=8pt,
    bottom=4pt,
    colback=gray!10!white,
    colframe=black,
    colbacktitle=black,
    enhanced,
    center,
    attach boxed title to top left={yshift=-0.1in,xshift=0.15in},
    boxed title style={boxrule=0pt,colframe=white,},
    fonttitle=\small\bfseries,
    before upper={\fontsize{8}{8}\selectfont},
  }
}
\newtcolorbox{AIbox}[2][]{aiboxstyle,title=#2,#1}

\tcbset{
  promptboxstyle/.style={
    width=\linewidth,
    top=8pt,
    bottom=4pt,
    colback=white!10!white,
    colframe=black,
    colbacktitle=black,
    enhanced,
    breakable,
    arc=3mm,
    frame hidden,
    borderline={1pt}{0pt}{black},
    skin first is subskin of={emptyfirst}{%
      frame code={\draw[black,line width=1pt,rounded corners=3mm] 
        (frame.south west) rectangle (frame.north east);}
    },
    skin middle is subskin of={emptymiddle}{%
      frame code={\draw[black,line width=1pt] 
        (frame.south west) rectangle (frame.north east);}
    },
    skin last is subskin of={emptylast}{%
      frame code={\draw[black,line width=1pt,rounded corners=3mm] 
        (frame.south west) rectangle (frame.north east);}
    },
    center,
    attach boxed title to top left={yshift=-0.1in,xshift=0.15in},
    boxed title style={boxrule=0pt,colframe=white,},
    fonttitle=\small\bfseries,
    before upper={\fontsize{10}{12}\selectfont},
  }
}
\newtcolorbox{Promptbox}[2][]{promptboxstyle,title=#2,#1}

\title{What's Missing in Screen-to-Action? Towards a UI-in-the-Loop Paradigm for Multimodal GUI Reasoning}

\author{
    Songze Li\textsuperscript{1,2},
    Xiaoke Guo\textsuperscript{1},
    Tianqi Liu\textsuperscript{1}, 
    Biao Yi\textsuperscript{1}, \\
    \textbf{Zhaoyan Gong}\textsuperscript{1,2},
    \textbf{Zhiqiang Liu}\textsuperscript{1},
    \textbf{Huajun Chen}\textsuperscript{1,2},
    \textbf{Wen Zhang\textsuperscript{1,2}\thanks{~~Corresponding authors.}}\\
    \textsuperscript{1}Zhejiang University,
    \textsuperscript{2}ZJU-Ant Group Joint Lab of Knowledge Graph \\
    \texttt{
    \{li.songze,zhang.wen\}@zju.edu.cn
    }
}

\begin{document}
\maketitle
\begin{abstract}
Existing Graphical User Interface (GUI) reasoning tasks remain challenging, particularly in UI understanding. Current methods typically rely on direct screen-based decision-making, which lacks interpretability and overlooks a comprehensive understanding of UI elements, ultimately leading to task failure. To enhance the understanding and interaction with UIs, we propose an innovative GUI reasoning paradigm called \textit{\textbf{UI-in-the-Loop}} (UILoop). Our approach treats the GUI reasoning task as a cyclic \textit{\textbf{Screen-UI elements-Action}} process. By enabling Multimodal Large Language Models (MLLMs) to explicitly learn the localization, semantic functions, and practical usage of key UI elements, UILoop achieves precise element discovery and performs interpretable reasoning. Furthermore, we introduce a more challenging \textit{\textbf{UI Comprehension}} task centered on UI elements with three evaluation metrics. Correspondingly, we contribute a benchmark of 26K samples (UI Comprehension-Bench) to comprehensively evaluate existing methods' mastery of UI elements. Extensive experiments demonstrate that UILoop achieves state-of-the-art UI understanding performance while yielding superior results in GUI reasoning tasks.\footnote{https://github.com/zjukg/UILoop}
\end{abstract}

\input{paper/section/introduction}

\input{paper/section/related_work}

\input{paper/section/methods}

\input{paper/section/experiments}

\input{paper/section/conclusion}

\section*{Limitations}

The primary limitations of our method encompass the following two aspects:

(1) UILoop primarily enhances the model's mastery of fine-grained UI elements but lacks consideration of UI layouts at different granularities within the screen, such as coarse-grained UI layouts composed of multiple fine-grained UI elements. In future work, we will further investigate the impact of UI elements at varying granularities on GUI reasoning capabilities.

(2) Current experiments predominantly focus on Qwen2.5-VL. In future work, we will explore the performance of UILoop across a broader range of MLLMs.

\section*{Ethics Statement}
In this paper, we introduce UI Comprehension-Bench, which is derived from existing GUI reasoning datasets Android Control, OmniAct, GUI-Act, ScreenSpot, ScreenSpot-Pro, and OS-Atlas, combined with externally collected webpages, mobile apps, and OS data. Furthermore, we conducted manual verification and excluded low-quality or non-compliant data, ensuring that our synthesized data does not violate any ethics. All UI screenshots were carefully reviewed to exclude or anonymize any personal or sensitive information. To promote transparency and reproducibility, we commit to releasing all code, models, and datasets upon publication of this paper, enabling the research community to verify our findings and build upon our work.

\section{Acknowledgement}

This work is founded by National Natural Science Foundation of China (NSFCU23B2055/NSFC62306276), New Generation Artificial Intelligence-National Science and Technology Major Project 2030 (2025ZD0122800), Yongjiang Talent Introduction Programme (2022A-238-G), and Fundamental Research Funds for the Central Universities (226-2023-00138). This work was supported by Ant Group. 


\bibliography{custom}

\newpage

\appendix

\input{paper/section/appendix}

\end{document}

%% file: paper/section/introduction.tex
\begin{figure*}[t]
    \centering
    \includegraphics[width=1.0\textwidth, keepaspectratio]{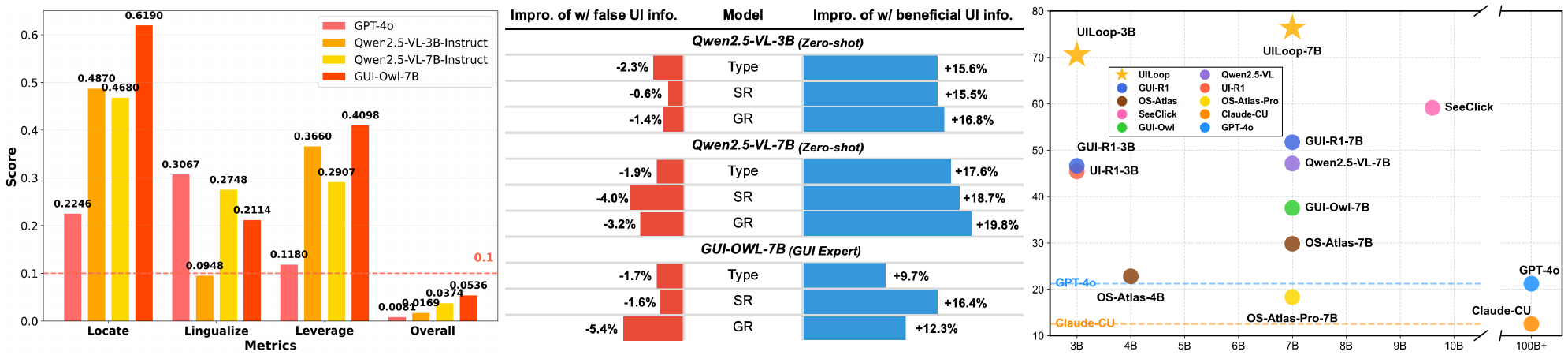} 
    \caption{\textbf{Left}: Evaluation of existing methods on UI element localization, semantic function description, and practical usage. \textbf{Middle}: Performance gains with correct vs. misleading UI info compared to without UI info. \textbf{Right}: Comparison of UILoop against existing \textit{\textbf{``Screen-to-Action"}} methods on SR metric for Android Control-High.}
    \label{fig:introduction-strength}
\end{figure*}

\section{Introduction}


GUI automation leverages Artificial Intelligence to simulate user interactions with device screens, reducing human workload \cite{nguyen-etal-2025-gui,shi2026androtmem}. Recent advances in MLLMs have significantly enhanced GUI agents \cite{Wang2023ASO,han2026unicorn,hu2026gui}, demonstrating substantial potential in web browsing, mobile apps, and office automation \cite{qin2025ui,hao2026recreate,lin2026mmfinereason,zhang2026expseek,ding2025arm,yu2026cirrusbench}, while advancing Artificial General Intelligence development \cite{ReTrack,hu2025agents,li2025cama,xiao2026staying,xiao2026layer,xiao2026promptbased,xiao2025visual,fu2025learning,shen2026fine}.

Existing GUI agents leverage advanced MLLMs (e.g., GPT-4o \cite{hurst2024gpt} and the Qwen-VL series \cite{bai2025qwen2}) to interpret user instructions and perform reasoning. However, these methods struggle with the complex layouts and diverse UI elements prevalent in real-world screens \cite{zhang2024large}. They typically follow a ``\textit{\textbf{Screen-Action}}" paradigm, where decisions and actions (e.g., click (123, 204), type ``text", scroll down) are generated directly from screen inputs \cite{wang-etal-2025-ponder, sun-etal-2025-os, Explorer, qi2024webrl}. This black-box decision-making process lacks interpretability and fails to foster a comprehensive understanding of UI elements \cite{wang2024gui}. 
Consequently, models often fail to accurately locate key elements and grasp their semantics and functions. Ultimately, this inability to effectively utilize these elements leads to task failure.

Evaluation of current GUI agents reveals significant deficiencies in UI element comprehension. As depicted in Fig. \ref{fig:introduction-strength} Left, advanced models exhibit poor performance (average score below 0.1) across three critical dimensions: UI element localization, semantic function description, and practical-usage. Based on this, we provide these models with both beneficial and misleading UI element descriptions during user instruction execution. Fig. \ref{fig:introduction-strength} Middle demonstrates that correct UI understanding substantially enhances reasoning across all scenarios—including zero-shot MLLMs, GUI expert, and models of varying scales. Conversely, incorrect descriptions significantly increase task failure rates. These findings underscore the critical role of UI element comprehension in GUI reasoning.

\begin{figure*}[ht]
    \centering
    \includegraphics[width=0.9\textwidth, keepaspectratio]{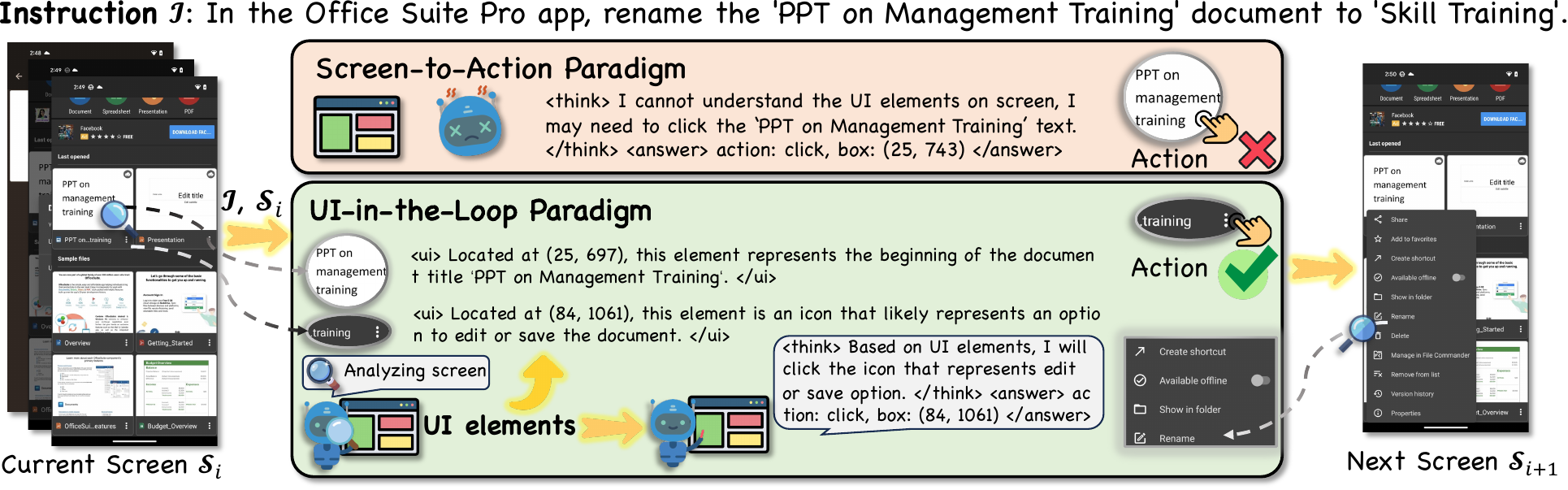} 
    \caption{Compared to the existing ``\textit{\textbf{Screen-to-Action}}" paradigm, our UI-in-the-Loop reframes GUI reasoning as ``\textit{\textbf{Screen-UI Elements-Action}}".}
    \label{fig:introduction-motivation}
\end{figure*}

To address the ``Missing in the Screen-to-Action" limitation inherent in current GUI models, we propose \textit{\textbf{UI-in-the-Loop}} (UILoop)—an innovative paradigm that reframes GUI reasoning around the mastery of UI elements. As illustrated in Fig. \ref{fig:introduction-motivation}, UILoop conceptualizes this process as a cyclic ``\textit{\textbf{Screen–UI Elements–Action}}" process, where UI elements serve as the critical bridge from screen to action, enabling more accurate reasoning based on correct UI elements. Leveraging reinforcement learning’s strength in handling complex sequential decisions \cite{shao2024deepseekmath,zhu2026medeyes,gong2026tempr1unifiedautonomousagent,lin2026medcausalx,zhu2026medsynapsevbridgingvisualperception,liu2026toolanchor,liu2026palm,shen2026fine,liu2025auditing}, we design UI‑Element‑Driven Reinforcement Fine‑Tuning, which teaches UILoop to locate key UI elements, infer their semantic functions, and master their practical usage, thereby achieving precise UI parsing and interpretable reasoning. Furthermore, recognizing the difficulty of understanding and applying UI elements, we introduce the more challenging \textit{\textbf{UI Comprehension}} task along with three evaluation metrics, and contribute a 26K benchmark (UI Comprehension-Bench) to comprehensively evaluate the UI localization, semantic understanding, and practical‑usage capabilities of existing models. Our major contributions are as follows:

\begin{itemize}[leftmargin=*]
    \item We demonstrate that comprehensive UI understanding significantly enhances reasoning in existing GUI agents. Building on this insight, we propose the innovative UILoop paradigm, which moves beyond conventional ``\textit{\textbf{Screen-to-Action}}" approaches by reframing GUI reasoning as cyclic ``\textit{\textbf{Screen–UI Elements–Action}}" loop. Through UI Element–Driven Reinforcement Fine-Tuning, UILoop improves model comprehension of interface elements, thereby advancing mutimodal GUI reasoning and interpretability.
    \item We introduce the more challenging \textit{\textbf{UI Comprehension}} task with three dedicated evaluation metrics (UI Locate, Lingualize, and Leverage) to assess how existing methods master UI elements. To support this, we advance community research by contributing UI Comprehension-Bench, a 26K benchmark for comprehensive UI capability assessment.
    \item Extensive experiments demonstrate that UILoop achieves state-of-the-art (SOTA) performance in UI comprehension, while delivering superior results in GUI reasoning tasks.
\end{itemize}

%% file: paper/section/related_work.tex
\section{Related Work}

\paragraph{Screen-to-Action GUI Agent.} Current approaches enhance GUI reasoning through large-scale pretraining (GUI-OWL \cite{ye2025mobileagentv3fundamentalagentsgui}) and supervised fine-tuning (Aguvis \cite{xu2024aguvis}, CoCo-Agent \cite{Ma2024CoCoAgentAC}, Show-UI \cite{11093438}, Aria-UI \cite{Aria-UI}). Moreover, recent work (UI-R1 \cite{lu2025ui}, GUI-R1 \cite{luo2025gui}, InfiGUI-R1 \cite{liu2025infigui}, InfiGUI-G1 \cite{liu2025infiguig1}, VeriOS \cite{wu2025verios}, VeriGUI \cite{zhang2026dontactblindlyrobust}) designs reinforcement learning for robust sequential decision-making \cite{lv2026pcsd,wu2026contrastive}. Several datasets such as Meta-GUI \cite{Sun2022METAGUITM}, AITW \cite{rawles2023androidinthewild}, GUIAct \cite{chen-etal-2025-guicourse}, OmniACT \cite{10.1007/978-3-031-73113-6_10}, Android Control \cite{Li2024OnTE}, AITZ \cite{zhang-etal-2024-android} have been proposed to enhance SFT or RL training for the ``Screen-to-Action" paradigm. However, this paradigm implicitly embeds UI comprehension within action prediction, lacking explicit UI element focus and limiting interpretability.

\paragraph{UI Elements-Enhanced GUI Agent.} Existing methods focus on UI element localization but ignore semantic functions and practical usage. SeeClick \cite{Cheng2024SeeClickHG} improves localization via ScreenSpot dataset. GUI-explorer \cite{xie-etal-2025-gui} retrieves UI information externally but doesn't enhance intrinsic understanding. ScreenSpot-Pro \cite{li2025screenspot}, MMBench-GUI \cite{wang2025mmbench}, UI-E2I-Bench \cite{liu2025ui}, UI-Vision \cite{nayak2025ui}, OS-Atlas \cite{wu2024atlas}, and UGround \cite{gou2025uground} improve localization but neglect their semantic and functional understanding, resulting in incorrect interactions such as clicking a scrollbar instead of dragging it. To address this, we propose UILoop, a ``Screen-UI Element-Action" paradigm that explicitly teaches models to master UI elements, achieving superior GUI reasoning performance.

%% file: paper/section/methods.tex
\section{Preliminary}


\paragraph{GUI Reasoning.} Given a user instruction $\mathcal{I}$, we formulate the GUI reasoning task as a multi-turn iterative decision-making process. At each step, the GUI Agent needs to interact with the current screen $\mathcal{S}_i$ and output an action. Therefore, our objective is to train the policy model $\pi_{\theta}$ to output the correct action $a_{i}$ to complete the user instruction:

\begin{center}
\scalebox{0.9}{$\displaystyle
\theta^{*} = {\underset{\theta}{\mathit{argmax}}{\prod\limits_{i}{P_{\pi_{\theta}}\left( a_{i} \middle| \mathcal{I},\mathcal{S}_{i} \right)}}},
$}
\end{center}
where $i$ is the $i$-th iteration cycle. Meanwhile, $\pi_{\theta}$ needs to analyze the UI elements in $\mathcal{S}_{i}$ that are beneficial for task completion: $\mathcal{U} = \left\{ u_{i} = \left\lbrack u_{i}^{loc} \in \mathcal{U}^{loc},u_{i}^{lin} \in \mathcal{U}^{lin},u_{i}^{lev} \in \mathcal{U}^{lev} \right\rbrack \right\}$, where $\mathcal{U}^{loc}, \mathcal{U}^{lin}, \mathcal{U}^{lev}$ represent location (e.g., (84, 1061)), semantic and functional description (e.g., ``this element is an icon that likely represents an option to edit or save the document"), and usage (e.g., action: click, box: (84, 1061)), respectively. By using $\mathcal{U}$ to obtain $a_{i}$, we can therefore model the objective as a ``\textit{\textbf{Screen–UI Elements–Action}}" iteration loop as follows:

\begin{center}
\scalebox{0.9}{$\displaystyle
\theta^{*} = {\underset{\theta}{\mathit{argmax}}{\prod\limits_{i}{P_{\pi_{\theta}}\left( a_{i} \middle| \mathcal{I},u_{j} \right)~{\prod\limits_{j}{P_{\pi_{\theta}}\left( u_{j} \middle| \mathcal{I},\mathcal{S}_{i} \right)}}}}}
$}
\end{center}

\paragraph{Group Relative Policy Optimization (GRPO)} \cite{guo2025deepseek} is a reinforcement learning algorithm for training models to improve performance on complex sequential decision-making (e.g., GUI reasoning) \citep{xie2026unlocking,xie2026edgeexperiencedistillationguidedexploration}. We employ GRPO to optimize our model. GRPO estimates the relative advantage of each response within a group of responses to the same prompt, eliminating the need for a value function. The optimization objective is:

\vspace{10pt}
\scalebox{0.9}{$\displaystyle
\begin{aligned}
\mathcal{L}(\theta) 
&= \mathbb{E}_{\{o_{i}\}_{i=1}^{G}\sim\pi_{\theta_{\text{old}}}(O|\mathcal{I}, \mathcal{S})} \\
&= \frac{1}{G}\sum_{i=1}^{G} \frac{1}{|o_{i}|} \sum_{t=1}^{|o_{i}|} 
\Bigg\{
\min\bigg[
\frac{\pi_{\theta}(o_{i,t}|\mathcal{I},o_{i,<t})}{\pi_{\theta_{\text{old}}}(o_{i,t}|\mathcal{I},o_{i,<t})} \\
&A_{i,t}^{\mathcal{U}}, 
\text{clip}\bigg(
\frac{\pi_{\theta}(o_{i,t}|\mathcal{I},o_{i,<t})}{\pi_{\theta_{\text{old}}}(o_{i,t}|\mathcal{I},o_{i,<t})}, 1-\epsilon, 1+\epsilon
\bigg) A_{i,t}^{\mathcal{U}}
\bigg] \\
&- \beta\,\mathbb{D}_{\text{KL}}(\pi_{\theta} \| \pi_{\text{ref}})
\Bigg\},
\end{aligned}
$}
where $G$ is the number of responses per $\mathcal{I}$, $o_i$ is the $i$-th response, $\pi_{\theta_{old}}$ is the old policy, $\pi_{\theta}$ is the current policy, $A_{i,t}^{\mathcal{U}}$ is the UI advantage of the $i$-th response at position $t$, $\epsilon$ is the clipping range, and $\mathbb{D}_{KL}\left( \pi_{\theta} \middle| \middle| \pi_{ref} \right)$ denotes the KL divergence penalty.

\section{UI-in-the-Loop Framework}

\begin{figure*}[ht]
    \centering
    \includegraphics[width=0.8\textwidth, keepaspectratio]{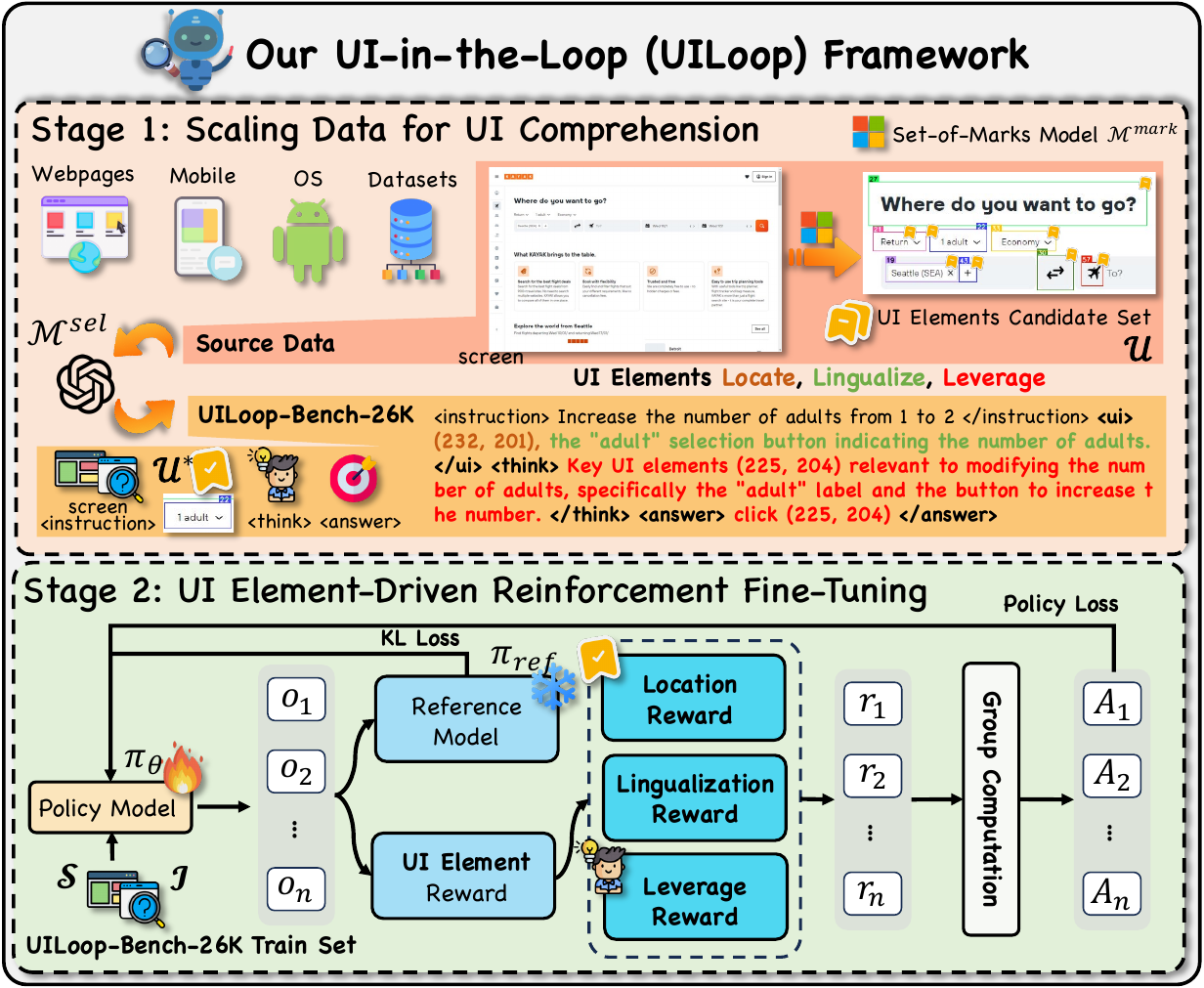} 
    \caption{Overview of our \textit{\textbf{UI-in-the-Loop}} (UILoop) framework.}
    \label{fig:method-framework}
\end{figure*}

As shown in Fig. \ref{fig:method-framework}, our GUI reasoning paradigm, \textit{\textbf{UI-in-the-Loop}} (UILoop), consists of two main stages. In the first stage, we design a Scaling Data for UI Comprehension synthesis pipeline to construct the UI Comprehension-Bench, serving to enhance the model's ability to understand and utilize UI elements. In the second stage, with this benchmark, we propose UI Element-Driven Reinforcement Fine-Tuning to address the ``Missing in the Screen-to-Action" limitation of existing models and strengthen the model's UI comprehension capabilities.

\subsection{Scaling Data for UI Comprehension}

\paragraph{Data Collection.} Existing GUI Reasoning datasets serve the ``Screen-to-Action" paradigm. Therefore, they lack fine-grained information regarding the location, semantic functionality, and practical usage of key UI elements on the screen. 
Consequently, we conduct a comprehensive augmentation of UI element information for existing GUI reasoning datasets.

Specifically, we collect training and testing data from Android Control \cite{Li2024OnTE}, OmniAct \cite{10.1007/978-3-031-73113-6_10}, GUI-Act \cite{chen-etal-2025-guicourse}, ScreenSpot \cite{Cheng2024SeeClickHG}, ScreenSpot-Pro \cite{li2025screenspot}, and OS-Atlas \cite{wu2024atlas} as source data, whose original data format is presented as ($\mathcal{I},\mathcal{S},a$). Based on this, we apply the set-of-marks model $\mathcal{M}^{mark}$ to $\mathcal{S}$ (e.g., OmniParser V2 \cite{yu2025omniparser}) to mark the locations of all identifiable UI elements as follows:

\begin{center}
\scalebox{0.9}{$\displaystyle
\left. \mathcal{M}^{mark}\left( \mathcal{S} \right)\rightarrow\mathcal{U}^{loc} \right.
$}
\end{center}

We employ GPT-4o as the selection model $\mathcal{M}^{sel}$ to filter out key UI elements that are beneficial for completing user instruction $\mathcal{I}$, and supplement the semantic functionality of these UI elements (as shown in Fig. \ref{fig:introduction-motivation}, included in <ui> along with location information) and practical usage (in the <think> and <answer> parts) described as follows:

\begin{center}
\scalebox{0.9}{$\displaystyle
\left. \mathcal{M}^{sel}\left( \mathcal{I},\mathcal{S},\mathcal{U}^{loc},a \right)\rightarrow\mathcal{U}^{*} \right.,
$}
\end{center}
where $\mathcal{U}^{*}$ represents the key UI elements. In addition, we perform fine-grained augmentation of UI element information for the dataset based on three different sources: Webpages, Mobile, and Operating System, following the same procedure as described above. Construction details are provided in the Appendix \ref{Details of UI Comprehension-Bench Collection}. Finally, we augment the fine-grained UI information and construct UI Comprehension-Bench, with data format: ($\mathcal{I},\mathcal{S},\mathcal{U}^{*},a$). Details are in Appendix \ref{Demonstrations of UI Comprehension-Bench}.


\paragraph{More than Action Prediction: UI Comprehension.} Existing GUI reasoning methods focus solely on ``Screen-to-Action" prediction, leaving the reasoning process a black box. Even when models output reasoning traces, they lack explicit modeling and evaluation of intermediate steps. Current evaluations measure only final action accuracy, neglecting UI element understanding and utilization, thus lacking interpretability. To address this, we propose a novel task: \textit{\textbf{UI Comprehension}}, which provides interpretable intermediate representations based on UI elements, establishing a transparent ``Screen-UI Element-Action" reasoning paradigm.

We design three evaluation metrics: Locate, Lingualize, and Leverage, assessing UI element localization, semantic function understanding, and utilization accuracy, respectively. The calculation of metrics is detailed in Sec. \ref{UI Element-Driven Reinforcement Fine-Tuning}. We define the final score as: Overall = Locate $*$ Lingualize $*$ Leverage. Furthermore, we contribute UI Comprehension-Bench 26K for this task.

\begin{figure*}[ht]
    \centering
    \includegraphics[width=1.0\textwidth, keepaspectratio]{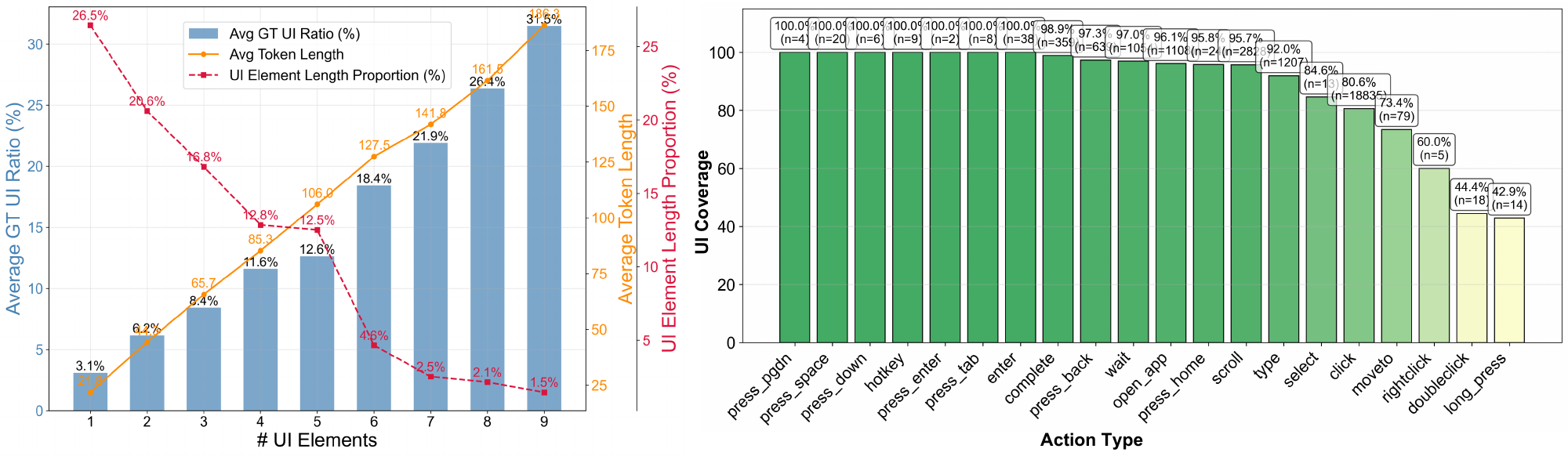} 
    \caption{Statistics of Our UI Comprehension-Bench. \textbf{Left:} Proportion and distribution of GT UI elements; token length of their semantic descriptions. \textbf{Right:} Proportion of GT UI elements effectively used in action inference.}
    \label{fig:method-statistics}
\end{figure*}

\label{Statistics of UI Comprehension-Bench}
\paragraph{Statistics of UI Comprehension-Bench.} Tab. \ref{tab:benchmark-statistic} compares our large-scale 26K UI Comprehension-Bench with existing GUI reasoning datasets. We are the first to provide Ground Truth (GT) UI elements (i.e., key UI elements) for screens and offer a fully interpretable ``Screen-UI Elements-Action" reasoning chain: locating GT UI elements, describing their semantic functions and practical usage, and finally deriving the action.

\input{paper/table/benchmark-statistic}

Fig. \ref{fig:method-statistics} presents detailed statistics. The benchmark contains 1,576,068 UI elements, with only 57,332 GT UI elements (<4\%), demonstrating identification difficulty. Fig. \ref{fig:method-statistics} Left visualizes the distribution of GT UI element proportions. When only 1 GT UI element exists, it comprises merely 3.1\% of total elements, requiring models to identify it among numerous irrelevant layouts. Such samples constitute 26.5\% of UI Comprehension-Bench, highlighting the difficulty. To verify our UI element effectiveness, we visualize text coverage rates of GT UI elements during reasoning, grouped by action type. Fig. \ref{fig:method-statistics} Right show coverage rates exceeding 90\% for most action types, with only minimal actions below 80\% (e.g., long\_press with 14 samples). This demonstrates that UI Comprehension-Bench provides high-quality UI elements with logical coherence and interpretability.

\subsection{UI Element-Driven Reinforcement Fine-Tuning}
\label{UI Element-Driven Reinforcement Fine-Tuning}

To address the ``Missing in the Screen-to-Action" limitation, we leverage reinforcement learning's strength in handling complex sequential decisions and propose UI Element-Driven Reinforcement Fine-Tuning to enhance the model's UI Comprehension capability. Specifically, we design Location, Lingualization, and Leverage Rewards to respectively strengthen the model's ability to locate UI elements, understand their semantic functions, and utilize them effectively. Firstly, we employ Format Reward to encourage the model to output in the expected format.

\paragraph{Format Reward.} We require the model to output in the following format.
\begin{AIbox}{Format}
<ui> Located at [x, y], describe the UI element’s semantics and function. </ui> <think> ... </think> <answer> [\{`action': , `point': , `input\_text': \}] </answer>
\end{AIbox}

If the output matches the expected format, the format reward is 1; otherwise, it is 0. 

\paragraph{Location Reward.} We use the Euclidean distance between the predicted UI element coordinates and the ground truth UI element coordinates as the location reward, defined as follows:

\vspace{3pt}
\scalebox{0.8}{$\displaystyle
\begin{aligned}
r^{loc} 
&= \frac{1}{|\mathcal{U}^{*}|} \sum_{i=1}^{|\mathcal{U}^{*}|}  1_{D}(u_{j}^{pred}) *[1-\\
&\quad \frac{\sqrt{(u_{i}^{loc^{*}}[x] - u_{j}^{loc^{pred}}[x])^{2} + (u_{i}^{loc^{*}}[y] - u_{j}^{loc^{pred}}[y])^{2}}}{\sqrt{w^{2} + h^{2}}}],
\end{aligned}
$}
where $w$ and $h$ denote the width and height of the screen, respectively, and $1_{D}(\cdot)$ is an indicator function that equals 1 when $u^{pred}$ is the nearest predicted UI element to $u^{*}$, and 0 otherwise.

\paragraph{Lingualization Reward.} We calculate the semantic similarity between the text descriptions of the predicted UI elements and the ground truth UI elements as follows:

\vspace{5pt}
\scalebox{0.8}{$\displaystyle
r^{lin} = \frac{1}{\left| \mathcal{U}^{*} \right|}{\sum\limits_{i = 1}^{|\mathcal{U}^{*}|}{1_{D}\left( u_{j}^{pred} \right)~*sim\left( u_{i}^{{lin}^{*}},u_{j}^{{lin}^{pred}} \right)}}
$}

\paragraph{Leverage Reward.} We adopt different calculation methods for action types in UI element utilization as follows. When the action type is `click':

\vspace{5pt}
\scalebox{0.8}{$\displaystyle
r^{lev} = 1_{A}\left( u_{j}^{{lev}^{pred}} \right)\left( u_{j}^{l{ev}^{pred}}\lbrack point\rbrack = = u^{{lev}^{*}}\lbrack point\rbrack \right)
$}
\vspace{5pt}

When the action type is one of `scroll', `type', `open\_app', or `select':

\vspace{5pt}
\scalebox{0.8}{$\displaystyle
r^{lev} = 1_{A}\left( u_{j}^{{lev}^{pred}} \right)\left( u_{j}^{l{ev}^{pred}}\lbrack text\rbrack = = u^{{lev}^{*}}\lbrack text\rbrack \right)
$}
\vspace{5pt}

For other actions, $r^{lev}=1_{A}\left( u_{j}^{{lev}^{pred}} \right)$. Here, $1_{A}(\cdot)$ is an indicator function that equals 1 when the action type of $u_{j}^{{lev}^{pred}}$ matches that of $u_{j}^{{lev}^{*}}$, and 0 otherwise. \textbf{We specifically note that the Location, Lingualize, and Leverage evaluation metrics of UI Comprehension-Bench are consistent with the calculation methods of the Location, Lingualization, and Leverage Rewards described above.} We define the overall reward as follows:

\begin{center}
\scalebox{0.8}{$\displaystyle
r~ = ~r^{format} + \alpha_{1}*r^{loc}*r^{lin} + \alpha_{2}*1_{U}\left( r^{loc}*r^{lin} \right)*{~r}^{lev}
$}
\end{center}

$1_{U}(\cdot)$ is an indicator function that equals 1 when $r^{loc}*r^{lin} > \eta$, and 0 otherwise. This design ensures that during training, the model \textbf{prioritizes locating key UI elements on the screen and understanding their semantic functions, and then learns to utilize these elements for accurate decision-making}.

Finally, we compute the advantage function using the obtained rewards as follows:

\vspace{5pt}
\begin{center}
\scalebox{0.8}{$\displaystyle
A_{i}^{\mathcal{U}} = \frac{r_{i} - mean\left( \left\{ r_{1},r_{2},...,r_{G} \right\} \right)}{std\left( \left\{ r_{1},r_{2},...,r_{G} \right\} \right)}
$}
\end{center}
where $mean$ and $std$ denote the mean and standard deviation, respectively.

%% file: paper/table/benchmark-statistic.tex
\begin{table*}[ht]
\centering
\resizebox{\textwidth}{!}{%
\begin{tabular}{lccccccccc}
\hline
\multicolumn{1}{c}{\multirow{3}{*}{\textbf{Datasets}}} & \multirow{3}{*}{\textbf{\# Episodes}} & \multirow{3}{*}{\textbf{\makecell{\# Unique \\ Instructions}}} & \multicolumn{7}{c}{\textbf{Annotation}} \\ \cline{4-10} 
\multicolumn{1}{c}{} &  &  & \multirow{2}{*}{\textbf{\makecell{Screen \\ Desc.}}} & \multicolumn{3}{c}{\textbf{Key UI Element}} & \multirow{2}{*}{\textbf{\makecell{Action \\ Coord}}} & \multirow{2}{*}{\textbf{\makecell{Action \\ Desc.}}} & \multirow{2}{*}{\textbf{\makecell{Action \\ Think}}} \\
\multicolumn{1}{c}{} &  &  &  & \textbf{Loc.} & \textbf{Lin.} & \textbf{Lev.} &  &  &  \\ \hline
AITW & 715142 & 30378 & \textcolor{red}{\ding{55}} & \textcolor{red}{\ding{55}} & \textcolor{red}{\ding{55}} & \textcolor{red}{\ding{55}} & \textcolor{green}{\ding{51}} & \textcolor{red}{\ding{55}} & \textcolor{red}{\ding{55}} \\
Android Control & 15283 & 15283 & \textcolor{green}{\ding{51}} & \textcolor{red}{\ding{55}} & \textcolor{red}{\ding{55}} & \textcolor{red}{\ding{55}} & \textcolor{green}{\ding{51}} & \textcolor{green}{\ding{51}} & \textcolor{red}{\ding{55}} \\
MMBench-GUI & 8123 & 8123 & \textcolor{red}{\ding{55}} & \textcolor{green}{\ding{51}} & \textcolor{red}{\ding{55}} & \textcolor{red}{\ding{55}} & \textcolor{green}{\ding{51}} & \textcolor{green}{\ding{51}} & \textcolor{green}{\ding{51}} \\
ScreenSpot-Pro & 1581 & 1581 & \textcolor{red}{\ding{55}} & \textcolor{green}{\ding{51}} & \textcolor{red}{\ding{55}} & \textcolor{red}{\ding{55}} & \textcolor{red}{\ding{55}} & \textcolor{red}{\ding{55}} & \textcolor{red}{\ding{55}} \\
UI-E2I-Bench & 1477 & 1477 & \textcolor{red}{\ding{55}} & \textcolor{green}{\ding{51}} & \textcolor{red}{\ding{55}} & \textcolor{red}{\ding{55}} & \textcolor{red}{\ding{55}} & \textcolor{red}{\ding{55}} & \textcolor{red}{\ding{55}} \\
UI-Vision & 8227 & $\sim$450 & \textcolor{green}{\ding{51}} & \textcolor{green}{\ding{51}} & \textcolor{red}{\ding{55}} & \textcolor{red}{\ding{55}} & \textcolor{green}{\ding{51}} & \textcolor{green}{\ding{51}} & \textcolor{red}{\ding{55}} \\ \hline
\textbf{Ours} & 26207 & 15735 & \textcolor{green}{\ding{52}} & \textcolor{green}{\ding{52}} & \textcolor{green}{\ding{52}} & \textcolor{green}{\ding{52}} & \textcolor{green}{\ding{52}} & \textcolor{green}{\ding{52}} & \textcolor{green}{\ding{52}} \\ \hline
\end{tabular}
}
\caption{Comparison of our UI Comprehension-Bench with existing GUI reasoning benchmarks.}
\label{tab:benchmark-statistic}
\end{table*}

%% file: paper/section/experiments.tex
\section{Experiments}

\subsection{Experiment Setting}
\paragraph{Datasets.} We evaluate on the test splits of Android Control-High and ScreenSpot-Pro, which assess high-difficulty multi-step GUI reasoning and cross-platform grounding, respectively. For UI Comprehension, we use UI Comprehension-Bench 26K, with statistics reported in Appendix \ref{Demonstrations of UI Comprehension-Bench}.

\paragraph{Evaluation Metrics.} 
We use action type accuracyw (Type), point accuracy (Ground Rate, GR), and step success rate (SR). Type measures action accuracy, GR assesses grounding capability, and SR evaluates overall accuracy of actions, coordinates, and text. For ScreenSpot-Pro, we use GR. For UI Comprehension, we use Locate, Lingualization, and Leverage to assess UI element grounding, semantic understanding, and utilization accuracy.

\paragraph{Baselines.} We compare: (1) Zero-shot general MLLMs performing GUI reasoning without training; (2) Screen-to-Action models—trained on GUI datasets to directly output actions from screens.

\paragraph{Implementation Details.} We use Qwen2.5-VL-3B and 7B as base models, trained on UI Comprehension-Bench's training set (Details in  Appendix \ref{Demonstrations of UI Comprehension-Bench}). We perform RFT using Verl \cite{sheng2024hybridflow} until reward convergence (3$\sim$6 epochs) with 5 rollouts. Prompts are detailed in the Appendix \ref{Prompt Details}. All experiments run on 8 A100 80G GPUs. $\alpha_{1}$ and $\alpha_{2}$ are set to 4, 5 separately. The UI indicator threshold $\eta$ is 0.5.

\subsection{Main Result}

\input{paper/table/main-results}

As shown in Tab. \ref{tab:main-results}, zero-shot MLLMs generally underperform training-based MLLMs due to lack of GUI training. Our method surpasses ``Screen-to-Action" models on both datasets. On ScreenSpot-Pro, our 3B and 7B models outperform similarly-sized Qwen2.5-VL and GUI-R1 by 13.3\%, 2\% and 13.3\%, 3.2\% in overall scores, respectively. On Android Control-High, our 7B model exceeds GUI expert models OS-Atlas-7B, OS-Atlas-Pro-7B, and GUI-OWL-7B by 46.5\%, 58.0\%, and 38.8\% in SR, respectively. These results demonstrate the superiority of the ``Screen-UI Element-Action" paradigm.

\subsection{Ablation Study}

\begin{figure*}[htbp]
    \centering
    \includegraphics[width=1.0\textwidth, keepaspectratio]{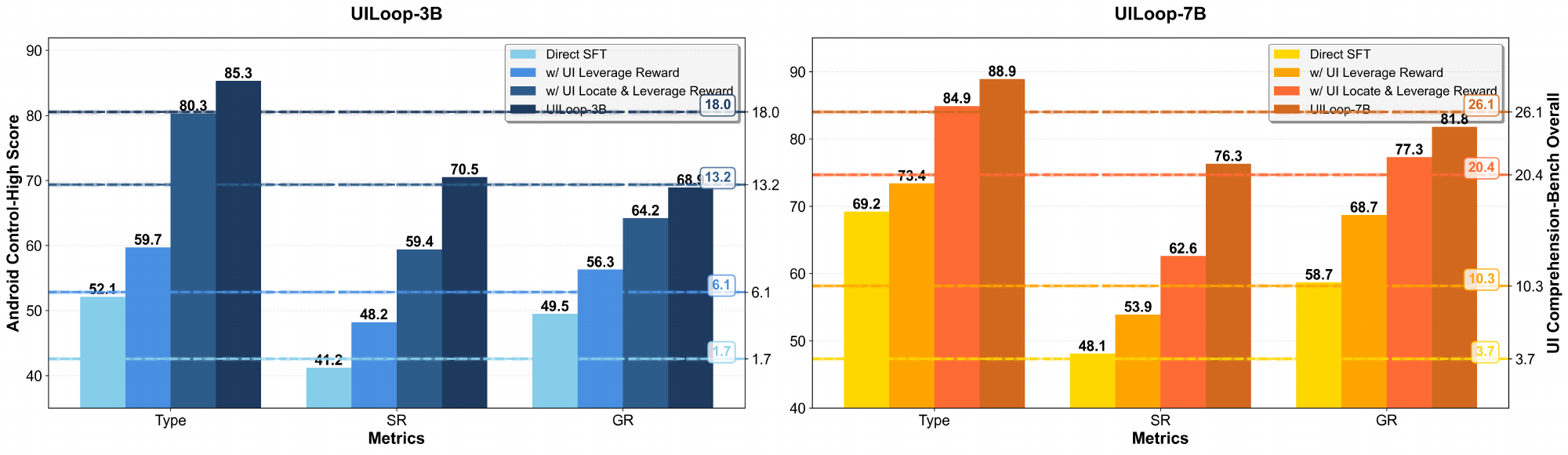} 
    \caption{Ablation Study on Android Control-High and UI Comprehension-Bench. We demonstrate the individual contributions of the Locate, Lingualize, Leverage Rewards on reasoning performance and UI comprehension.}
    \label{fig:experiment-ablation}
\end{figure*}

\input{paper/table/impact-ui-element}

We conducted ablation studies to examine the impact of different UI Rewards on reasoning performance, as shown in Fig. \ref{fig:experiment-ablation}. We evaluated: (1) Direct SFT; (2) Direct RFT with Leverage Reward only; (3) Locate + Leverage Rewards; (4) Full UILoop. Results show that Leverage Reward improves all metrics by teaching models to analyze and utilize UI elements. Adding Locate Reward increases GR by 7.9\% and 8.6\% for 3B and 7B models, enhancing key UI element localization and action positioning accuracy. Further adding Lingualize Reward improves SR by 11.1\% and 13.7\%, strengthening semantic understanding of key UI elements and action text accuracy. These results validate that each reward effectively enhances reasoning by improving UI element mastery.

\subsection{Impact of UI Elements}

As shown in Tab. \ref{tab:impact-ui-element}, we examined three UI intervention approaches: (1) key UI element info., (2) false UI element info., and (3) UILoop Training. Results show false UI info. impairs GUI reasoning, while key UI info. as context significantly improves accuracy, demonstrating that enhancing key UI mastery benefits GUI reasoning. Moreover, UILoop Training surpasses merely providing key UI info., achieving improvements of 31.6\% and 22.8\% on Qwen2.5-VL-3B and 7B (versus 16.0\% and 18.7\%), and 17.8\% and 29.0\% on GUI-Owl-7B and OS-Atlas-Pro-7B (versus 12.8\% and 20.7\%) for context alone, demonstrating its superiority in enhancing intrinsic UI comprehension and reasoning performance.

\subsection{Experiment of UI Comprehension-Bench}

\begin{figure}[!htbp]
    \centering
    \includegraphics[width=1.0\linewidth, keepaspectratio]{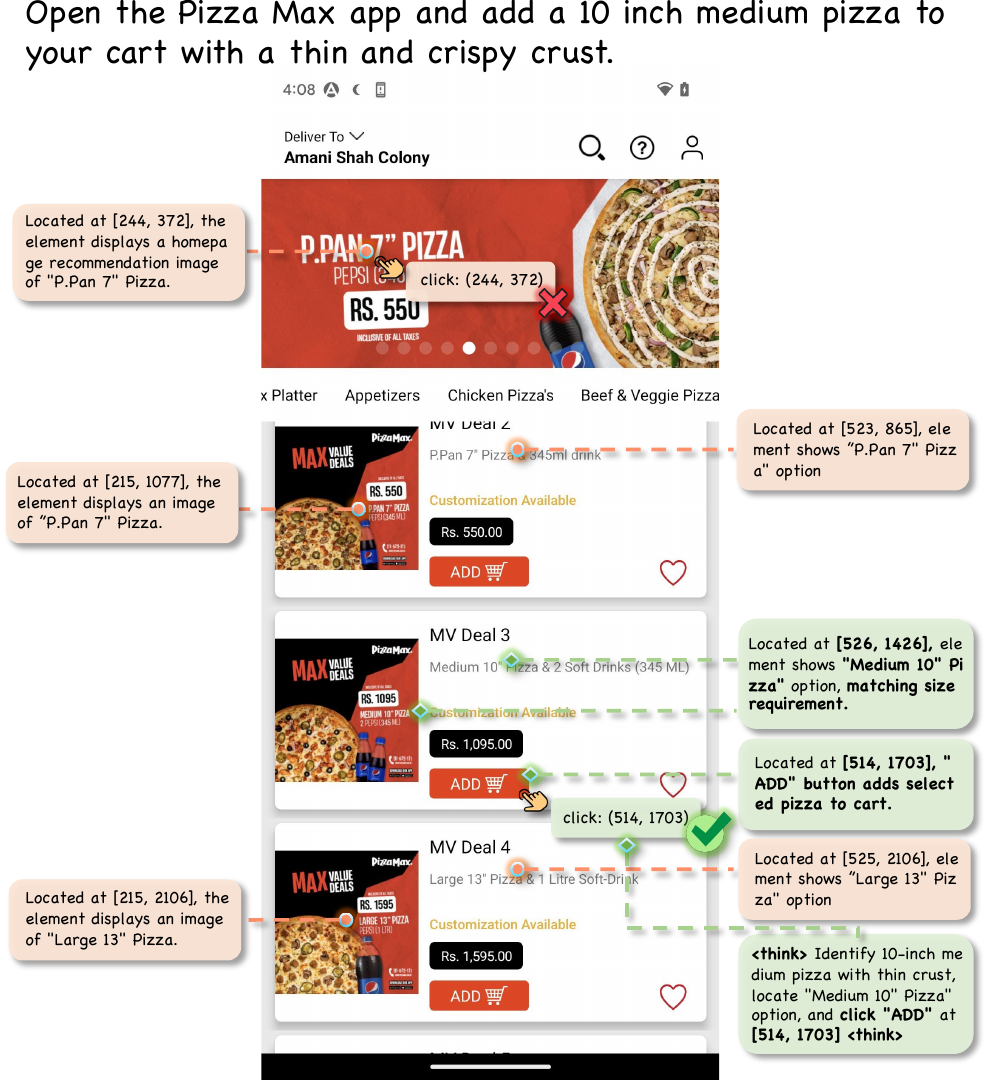} 
    \caption{Comparative Case Study between UILoop and ``Screen-to-Action".}
    \label{fig:experiment-case-study}
\end{figure}

\input{paper/table/ui-comprehension-bench}

We evaluated existing models on our UI Comprehension-Bench, as shown in Tab. \ref{tab:ui-comprehension-bench}. Results reveal that current ``Screen-to-Action" models perform poorly across Locate, Lingualize, and Leverage tasks, all scoring below 10\%. In contrast, UILoop achieves a SOTA score of 26.1 on the 7B model, and boosts the overall scores of GUI-Owl-7B and OS-Atlas-Pro-7B by 18.4 and 9.0 (underline parts), demonstrating its superiority in enhancing UI comprehension. Our UI Comprehension-Bench will advance GUI agents from ``Screen-to-Action" toward the more superior ``Screen-UI Element-Action" paradigm, providing the first robust benchmark for UI comprehension capabilities.

\subsection{Case Study}

We conducted a case study as shown in Fig. \ref{fig:experiment-case-study}. For the instruction ``Open the Pizza Max app and add a 10 inch medium pizza to your cart with a crust," key UI elements (\textcolor{green}{Green}) and misleading ones (\textcolor{red}{Red}) have minimal visual differences. ``Screen-to-Action" methods incorrectly click ``P. PAN 7", while UILoop correctly identifies ``Medium 10" by analyzing UI element semantics and the ``ADD" button's function. UILoop also explicitly shows the reasoning process from Screen to key UI elements to Action, demonstrating superior interpretability.


%% file: paper/table/main-results.tex

\begin{table*}[ht]
\centering
\resizebox{1.0\textwidth}{!}{%
\begin{tabular}{lcccccccccccccccccc}
\toprule
\multicolumn{1}{c}{\multirow{3}{*}{Methods}} & \multicolumn{15}{c}{\textit{\textbf{ScreenSpot-Pro}}} & \multicolumn{3}{c}{\textit{\textbf{AndroidControl-High}}} \\ \cline{2-19} 
\multicolumn{1}{c}{} & \multicolumn{2}{c}{Dev} & \multicolumn{2}{c}{Creative} & \multicolumn{2}{c}{CAD} & \multicolumn{2}{c}{Sci.} & \multicolumn{2}{c}{Office} & \multicolumn{2}{c}{OS} & \multicolumn{3}{c}{Overall} & \multirow{2}{*}{Type} & \multirow{2}{*}{SR} & \multirow{2}{*}{GR} \\
\multicolumn{1}{c}{} & Text & Icon & Text & Icon & Text & Icon & Text & Icon & Text & Icon & Text & Icon & Text & Icon & Avg. &  &  &  \\ \hline
\rowcolor{gray!20}
\multicolumn{19}{c}{\textit{\textbf{Zero-Shot   Models}}} \\ \hline
Claude-CU & 22.0 & 3.9 & 25.9 & 3.4 & 14.5 & 3.7 & 33.9 & 15.8 & 30.1 & 16.3 & 11.0 & 4.5 & 23.4 & 7.1 & \cellcolor{magenta!20}17.1 & 63.7 & \cellcolor{green!20}12.5 & - \\
GPT-4o & 1.3 & 0.0 & 1.0 & 0.0 & 2.0 & 0.0 & 2.1 & 0.0 & 1.1 & 0.0 & 0.0 & 0.0 & 1.3 & 0.0 & \cellcolor{magenta!20}0.8 & 63.1 & \cellcolor{green!20}21.2 & 30.9 \\
Qwen2.5-VL-3B & 16.2 & 1.4 & 23.3 & 1.4 & 10.2 & 4.7 & 38.2 & 6.4 & 24.3 & 3.8 & 15.0 & 1.1 & 21.2 & 3.1 & \cellcolor{magenta!20}12.2 & 47.8 & \cellcolor{green!20}38.9 & 46.5 \\
Qwen2.5-VL-7B & 33.1 & 2.1 & 23.7 & 3.5 & 12.2 & 6.3 & 36.8 & 7.3 & 37.8 & 7.5 & 30.8 & 6.9 & 29.1 & 5.6 & \cellcolor{magenta!20}17.4 & 68.7 & \cellcolor{green!20}47.1 & 59.7 \\ \hline
\rowcolor{gray!20}
\multicolumn{19}{c}{\textit{\textbf{Screen-to-Action   Training Models}}} \\ \hline
SeeClick & 0.6 & 0.0 & 1.0 & 0.0 & 2.5 & 0.0 & 3.5 & 0.0 & 1.1 & 0.0 & 2.8 & 0.0 & 1.8 & 0.0 & \cellcolor{magenta!20}1.1 & 82.9 & \cellcolor{green!20}59.1 & 62.9 \\
GUI-Owl-7B & 37.0 & 5.5 & 32.8 & 1.4 & 23.9 & 4.7 & 37.5 & 10.0 & 33.9 & 11.3 & 18.7 & 3.4 & 31.0 & 5.5 & \cellcolor{magenta!20}21.3 & 72.9 & \cellcolor{green!20}37.5 & 53.7 \\
OS-Atlas-Pro-7B & 1.4 & 0.0 & 1.1 & 0.0 & 2.7 & 0.0 & 1.5 & 0.0 & 1.8 & 2.0 & 0.0 & 0.0 & 1.4 & 0.3 & \cellcolor{magenta!20}0.9 & 69.7 & \cellcolor{green!20}18.3 & 16.8 \\
OS-Atlas-4B & 7.1 & 0.0 & 3.0 & 1.4 & 2.0 & 0.0 & 9.0 & 5.5 & 5.1 & 3.8 & 5.6 & 0.0 & 5.0 & 1.7 & \cellcolor{magenta!20}3.7 & 49.0 & \cellcolor{green!20}22.8 & 49.5 \\
OS-Atlas-7B & 33.1 & 1.4 & 28.8 & 2.8 & 12.2 & 4.7 & 37.5 & 7.3 & 33.9 & 5.7 & 27.1 & 4.5 & 28.1 & 4.0 & \cellcolor{magenta!20}18.9 & 57.4 & \cellcolor{green!20}29.8 & 54.9 \\
Qwen2.5-VL-3B$^{*}$ & 20.3 & 1.8 & 24.6 & 2.8 & 11.2 & 4.7 & 39.5 & 6.4 & 28.6 & 5.7 & 17.8 & 2.2 & 23.8 & 3.9 & \cellcolor{magenta!20}13.9 & 52.1 & \cellcolor{green!20}41.2 & 49.5 \\
Qwen2.5-VL-7B$^{*}$ & 31.4 & 1.8 & 27.3 & 3.5 & 15.7 & 5.1 & 40.7 & 7.9 & 39.7 & 8.9 & 32.4 & 6.9 & 31.2 & 5.7 & \cellcolor{magenta!20}18.5 & 69.2 & \cellcolor{green!20}48.1 & 58.7 \\
ShowUI-2B & 16.9 & 1.4 & 9.1 & 0.0 & 2.5 & 0.0 & 13.2 & 7.3 & 15.3 & 7.5 & 10.3 & 2.2 & 10.8 & 2.6 & \cellcolor{magenta!20}7.7 & - & \cellcolor{green!20}- & - \\
Aria-UI & 16.2 & 0.0 & 23.7 & 2.1 & 7.6 & 1.6 & 27.1 & 6.4 & 20.3 & 1.9 & 4.7 & 0.0 & 17.1 & 2.0 & \cellcolor{magenta!20}11.3 & - & \cellcolor{green!20}10.2 & 43.2 \\
UI-R1-3B & 22.7 & 4.1 & 27.3 & 3.5 & 11.2 & 6.3 & 43.4 & 11.8 & 32.2 & 11.3 & 13.1 & 4.5 & 25.0 & 6.9 & \cellcolor{magenta!20}17.8 & 57.9 & \cellcolor{green!20}45.4 & 55.7 \\
UGround-7B & 26.6 & 2.1 & 27.3 & 2.8 & 14.2 & 1.6 & 31.9 & 2.7 & 31.6 & 11.3 & 17.8 & 0.0 & 25.0 & 2.8 & \cellcolor{magenta!20}16.5 & - & \cellcolor{green!20}- & - \\
GUI-R1-3B & 33.8 & 4.8 & 40.9 & 5.6 & 26.4 & 7.8 & \underline{61.8} & \underline{17.3} & \underline{53.6} & \underline{17.0} & \underline{28.1} & 5.6 & 40.7 & \underline{9.7} & \cellcolor{magenta!20}25.2 & 58.0 & \cellcolor{green!20}46.6 & 56.2 \\
GUI-R1-7B & 49.4 & 4.8 & 38.9 & 8.4 & 23.9 & 6.3 & \textbf{55.6} & 11.8 & \textbf{58.7} & \textbf{26.4} & \textbf{42.1} & \textbf{16.9} & 44.8 & \textbf{12.4} & \cellcolor{magenta!20}28.6 & 71.6 & \cellcolor{green!20}51.7 & 65.6 \\ \hline
\rowcolor{gray!20}
\multicolumn{19}{c}{\textit{\textbf{UILoop Training Models}}} \\ \hline
UILoop-3B & \underline{46.1} & \underline{4.8} & \underline{45.6} & \underline{7.8} & \underline{32.5} & \underline{8.5} & 48.2 & 15.0 & 49.3 & 10.8 & 26.4 & \underline{7.7} & \underline{41.3} & 9.1 & \cellcolor{magenta!20}\underline{27.2} & \underline{85.3} & \cellcolor{green!20}\underline{70.5} & \underline{68.9} \\
UILoop-7B & \textbf{52.6} & \textbf{9.7} & \textbf{47.4} & \textbf{9.1} & \textbf{38.3} & \textbf{12.5} & 49.6 & \textbf{15.2} & 51.1 & 12.7 & 34.8 & 8.1 & \textbf{45.5} & 11.2 & \cellcolor{magenta!20}\textbf{31.8} & \textbf{88.9} & \cellcolor{green!20}\textbf{76.3} & \textbf{81.8} \\ \bottomrule
\end{tabular}
}
\caption{Performance comparison of UILoop with zero-shot and ``Screen-to-Action" paradigm models on ScreenSpot-Pro and AndroidControl-High. $^{*}$ denotes SFT models trained on \cite{luo2025gui}. \underline{Underline} and \textbf{bold} indicate the best results among 3B and 7B models, respectively.}
\label{tab:main-results}
\end{table*}

%% file: paper/table/impact-ui-element.tex
\begin{table}[ht]
\centering
\resizebox{0.43\textwidth}{!}{%
\begin{tabular}{lcccc}
\hline
\multicolumn{1}{c}{\multirow{2}{*}{Methods}} & \multicolumn{3}{c}{\textit{\textbf{Android Control-High}}} & Impact \\
\multicolumn{1}{c}{} & Type & SR & GR & Avg. Ratio \\ \hline
\rowcolor{gray!20}
\multicolumn{5}{c}{\textit{\textbf{GPT-4o-mini$_{\text{(Zero-shot)}}$}}} \\ \hline
base & 68.1 & 20.9 & 6.9 & - \\
w/ UI info. & 69.9 & 51.4 & 62.9 & +29.4 \\
w/ false UI info. & 67.2 & 18.4 & 5.8 & -1.5 \\ \hline
\rowcolor{gray!20}
\multicolumn{5}{c}{\textit{\textbf{Qwen2.5-VL-3B-Instruct$_{\text{(Zero-shot)}}$}}} \\ \hline
base & 58.2 & 32.7 & 39.0 & - \\
w/ UI info. & 73.8 & 48.3 & 55.8 & +16.0 \\
w/ false UI info. & 55.9 & 32.1 & 37.6 & -1.4 \\
\rowcolor{green!20}
w/ UILoop & 85.3 & 70.5 & 68.9 & \textbf{+31.6} \\ \hline
\rowcolor{gray!20}
\multicolumn{5}{c}{\textit{\textbf{Qwen2.5-VL-7B-Instruct$_{\text{(Zero-shot)}}$}}} \\ \hline
base & 68.3 & 53.6 & 56.7 & - \\
w/ UI info. & 86.0 & 72.3 & 76.5 & +18.7 \\
w/ false UI info. & 66.4 & 49.6 & 53.5 & -3.0 \\
\rowcolor{green!20}
w/ UILoop & 88.9 & 76.3 & 81.8 & \textbf{+22.8} \\ \hline
\rowcolor{gray!20}
\multicolumn{5}{c}{\textit{\textbf{GUI-Owl-7B$_{\text{(GUI Expert)}}$}}} \\ \hline
base & 72.9 & 37.5 & 53.7 & - \\
w/ UI info. & 82.6 & 53.8 & 66.1 & +12.8 \\
w/ false UI info. & 71.2 & 35.8 & 48.4 & -2.9 \\
\rowcolor{green!20}
w/ UILoop & 84.9 & 64.7 & 68.0 & \textbf{+17.8} \\ \hline
\rowcolor{gray!20}
\multicolumn{5}{c}{\textit{\textbf{OS-Atlas-Pro-7B$_{\text{(GUI Expert)}}$}}} \\ \hline
base & 69.7 & 18.3 & 16.8 & - \\
w/ UI info. & 73.3 & 45.1 & 48.5 & +20.7 \\
w/ false UI info. & 54.6 & 16.7 & 15.0 & -6.2 \\
\rowcolor{green!20}
w/ UILoop & 80.3 & 57.6 & 53.9 & \textbf{+29.0} \\ \hline
\end{tabular}
}
\caption{Impact of different UI element intervention methods on GUI reasoning performance.}
\label{tab:impact-ui-element}
\end{table}

%% file: paper/table/ui-comprehension-bench.tex
\begin{table}[ht]
\centering
\resizebox{0.44\textwidth}{!}{%
\begin{tabular}{lcccc}
\hline
\multicolumn{1}{c}{\multirow{2}{*}{Methods}} & \multicolumn{4}{c}{\textit{\textbf{UI Comprehension-Bench}}} \\
\multicolumn{1}{c}{} & Loc. & Lin. & Lev. & Overall \\ \hline
\rowcolor{gray!20}
\multicolumn{5}{c}{\textit{\textbf{Zero-shot Models}}} \\ \hline
GPT-4o & 22.5 & 30.7 & 11.8 & \cellcolor{green!20}0.8 \\
Qwen2.5-VL-3B-Instruct & 48.7 & 9.5 & 36.6 & \cellcolor{green!20}1.7 \\
Qwen2.5-VL-7B-Instruct & 46.8 & 27.5 & 29.1 & \cellcolor{green!20}3.7 \\ \hline
\rowcolor{gray!20}
\multicolumn{5}{c}{\textit{\textbf{Screen-to-Action Training Models}}} \\ \hline
GUI-Owl-7B & 61.9 & 21.1 & 41.0 & \cellcolor{green!20}5.4 \\
\quad w/ UILoop & 87.4 & 51.1 & 53.4 & \cellcolor{green!20}\underline{23.8} \\
OS-Atlas-Pro-7B & 49.6 & 48.2 & 18.9 & \cellcolor{green!20}4.5 \\
\quad w/ UILoop & 71.4 & 54.2 & 34.9 & \cellcolor{green!20}\underline{13.5} \\
UI-R1-3B & 47.1 & 39.7 & 33.7 & \cellcolor{green!20}6.3 \\
GUI-R1-3B & 47.4 & 37.9 & 35.9 & \cellcolor{green!20}6.4 \\
GUI-R1-7B & 62.6 & 47.6 & 35.3 & \cellcolor{green!20}10.5 \\
\rowcolor{gray!20}
\multicolumn{5}{c}{\textit{\textbf{UILoop Training Models}}} \\ \hline
UILoop-3B & 80.3 & 44.7 & 50.2 & \cellcolor{green!20}18.0 \\
UILoop-7B & 86.4 & 49.3 & 61.3 & \cellcolor{green!20}\textbf{26.1} \\ \hline
\end{tabular}
}
\caption{Overall performance of different paradigm methods on UI element Locate, Lingualize, and Leverage capabilities in our UI Comprehension-Bench.}
\label{tab:ui-comprehension-bench}
\end{table}

%% file: paper/section/conclusion.tex
\section{Conclusion}

In this paper, we highlight that comprehensive UI understanding significantly enhances GUI agent reasoning. We propose \textit{\textbf{UI-in-the-Loop}} (UILoop), an innovative paradigm that reframes GUI reasoning from conventional \textit{\textbf{``Screen-to-Action"}} to a cyclic \textit{\textbf{``Screen–UI Elements–Action"}} loop. We design UI Element-Driven Reinforcement Fine-Tuning to improve interface element comprehension, advancing multimodal GUI reasoning and interpretability.
To facilitate this research, we introduce the \textit{\textbf{UI Comprehension}} task with three evaluation metrics (UI Locate, Lingualize, and Leverage) and contribute UI Comprehension-Bench, a 26K benchmark for comprehensive UI assessment. Extensive experiments show UILoop achieves state-of-the-art performance in UI comprehension and delivers superior results in GUI reasoning tasks.

%% file: paper/section/appendix.tex
\section{Details of UI Comprehension-Bench Collection}
\label{Details of UI Comprehension-Bench Collection}


We elaborate on the data synthesis pipeline of UI Comprehension-Bench in this section. Our pipeline primarily consists of three steps: Source Data Collection, Key UI Element Identification and Parsing, and Human Verification.

\paragraph{Source Data Collection.} Our data sources mainly include webpages, mobile applications, operating systems, and existing GUI reasoning datasets. For webpages, we capture screens from real browsers using BrowserGym \cite{chezelles2024browsergym} and Playwright \footnote{https://github.com/microsoft/playwright}, randomly simulate actions such as clicking, scrolling, and typing on the screens, and retain successfully executed actions. For mobile and OS data, we employ DroidBot\footnote{https://github.com/honeynet/droidbot} \cite{wen2023droidbot} to perform the same screen capture and action execution procedures on real Android applications and operating systems. We also incorporate training data from existing datasets—Android Control, OmniAct, GUI-Act, ScreenSpot, ScreenSpot-Pro, and OS-Atlas—as part of our source data. We normalize the format of all source data, with each sample containing the following data fields: (instruction, screen, action).

\paragraph{Key UI Element Identification and Parsing.} We process the screens obtained from the source data by employing a set-of-marks model, specifically OmniParser V2, to annotate all identifiable UI elements on the screen. This enables us to obtain coordinate information for all candidate UI elements. We then utilize GPT-4o as a selection model to identify UI elements that are beneficial for completing the given instruction and to provide reasoning processes explaining how these UI elements contribute to task completion. Specifically, we input (instruction, screen, UI element coordinate information, action) into the selection model to identify key UI elements and generate their semantic functions and practical usage (detailed prompts are provided in Appendix \ref{Prompt Details}). Consequently, we expand the data format of the source data to (instruction, screen, key UI element information, action).

\paragraph{Human Verification.} We conduct manual screening of the obtained data to exclude samples with incorrect instructions, erroneous answers, or misidentified key UI elements. Through this verification process, \textbf{we ultimately curate UI Comprehension-Bench, which comprises 26,207 samples, including a training set of 3,471 samples (selected from the training sets of Android Control, OmniAct, GUI-Act, ScreenSpot, ScreenSpot-Pro, and OS-Atlas) and a test set of 22,736 samples, ensuring complete data isolation between the two sets.}

\section{Demonstrations of UI Comprehension-Bench}
\label{Demonstrations of UI Comprehension-Bench}

\input{paper/table/details-benchmark-statistic}

In this section, we compare UI Comprehension‑Bench with existing GUI reasoning datasets and present UI Comprehension‑Bench through detailed example instances. Existing GUI‑reasoning datasets (including PixelHelp \cite{li2020mapping}, MoTIF \cite{10.1007/978-3-031-20074-8_18}, UGIF \cite{gubbi-venkatesh-etal-2024-ugif}, Meta-GUI \cite{Sun2022METAGUITM}, AITW \cite{rawles2023androidinthewild}, GUIAct \cite{chen-etal-2025-guicourse}, OmniACT \cite{10.1007/978-3-031-73113-6_10}, Android Control \cite{Li2024OnTE}, AITZ \cite{zhang-etal-2024-android}, MMBench-GUI \cite{wang2025mmbench}, ScreenSpot \cite{Cheng2024SeeClickHG}, ScreenSpot-V2 \cite{Cheng2024SeeClickHG}, ScreenSpot-Pro \cite{li2025screenspot}, UI-E2I-Bench \cite{liu2025ui}, UI-Vision \cite{nayak2025ui}) follow the ``Screen‑to‑Action" paradigm \cite{dong2025aurora,dong2026neureasonerexplainablecontrollableunified,jiang2026foeforesterrorsmakes,liu2026general365,kwan2026videommelogicalcontrolleddiagnosticbenchmark,xiao2026reversible,zhang2026pi,zhang2025trimtokenator}. Consequently, they lack fine‑grained information about the location, semantic functionality, and practical usage of key UI elements on the screen, as shown in Tab. \ref{tab:details-benchmark-statistic}.

\begin{figure*}[!htbp]
    \centering
    \includegraphics[width=1.0\textwidth, keepaspectratio]{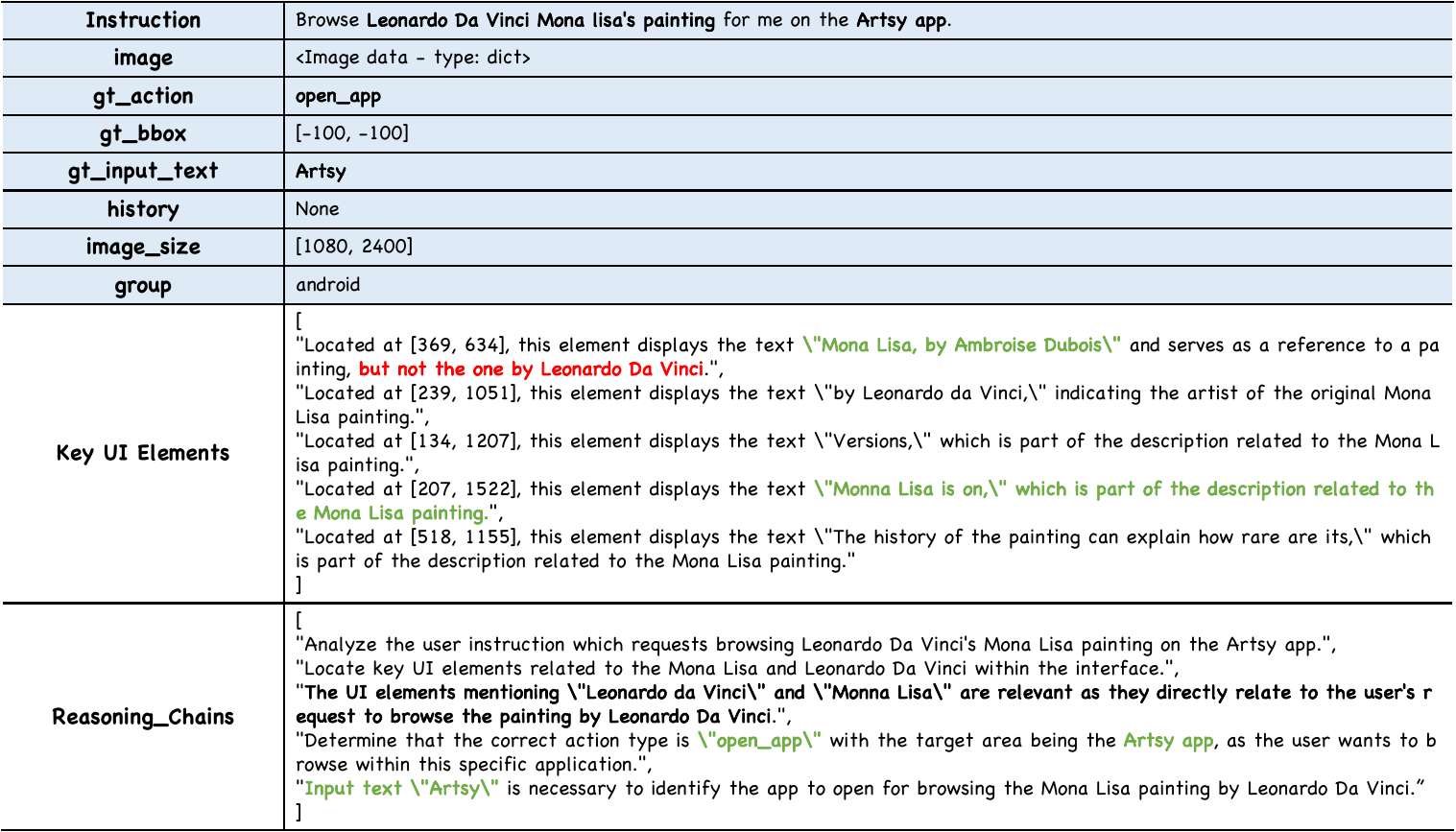} 
    \caption{Case with \textit{open\_app} actions in our UI Comprehension-Bench.}
    \label{fig:case-study1}
\end{figure*}

\begin{figure*}[!htbp]
    \centering
    \includegraphics[width=1.0\textwidth, keepaspectratio]{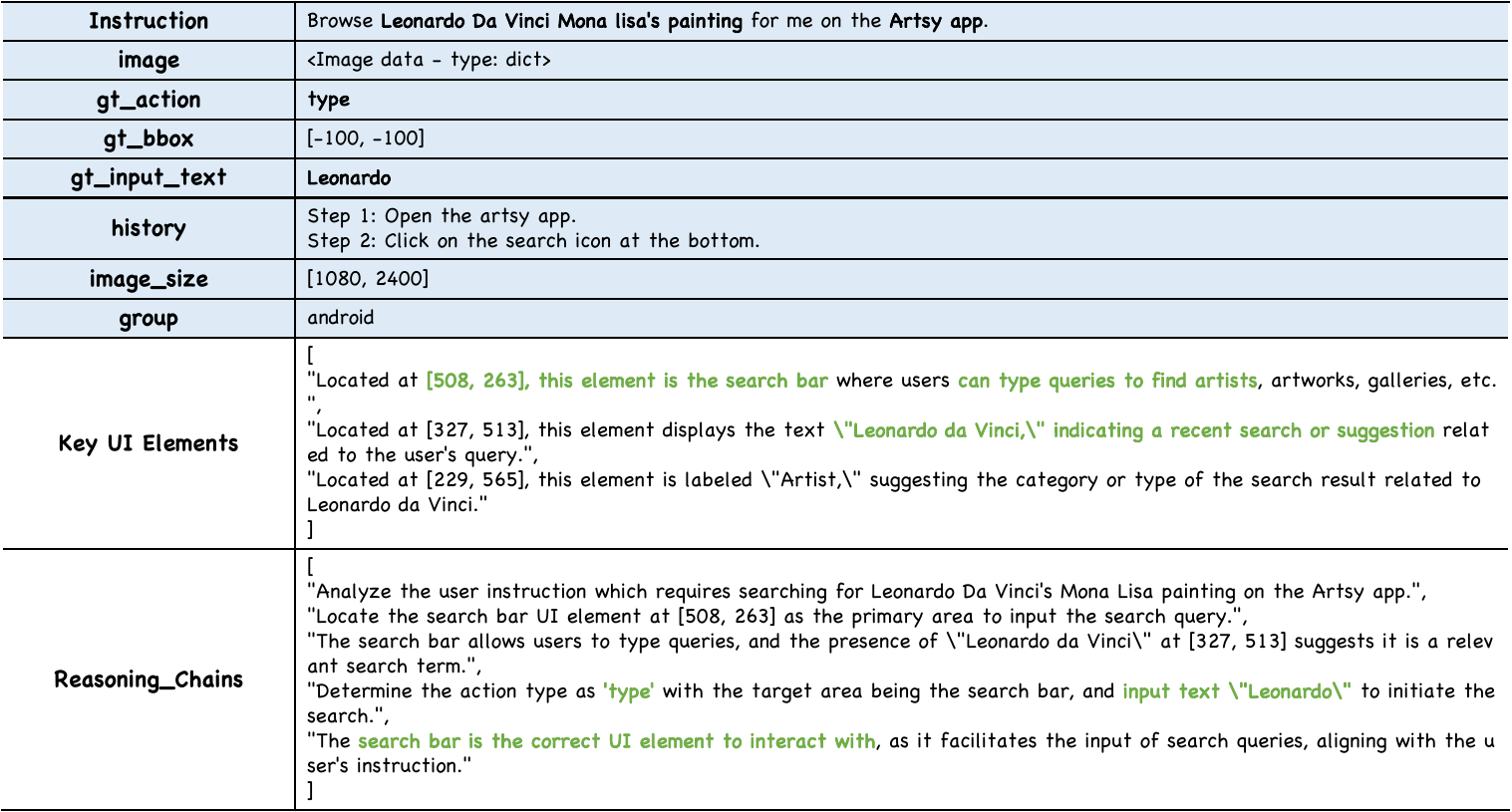} 
    \caption{Case with \textit{type} actions in our UI Comprehension-Bench.}
    \label{fig:case-study2}
\end{figure*}

\begin{figure*}[!htbp]
    \centering
    \includegraphics[width=1.0\textwidth, keepaspectratio]{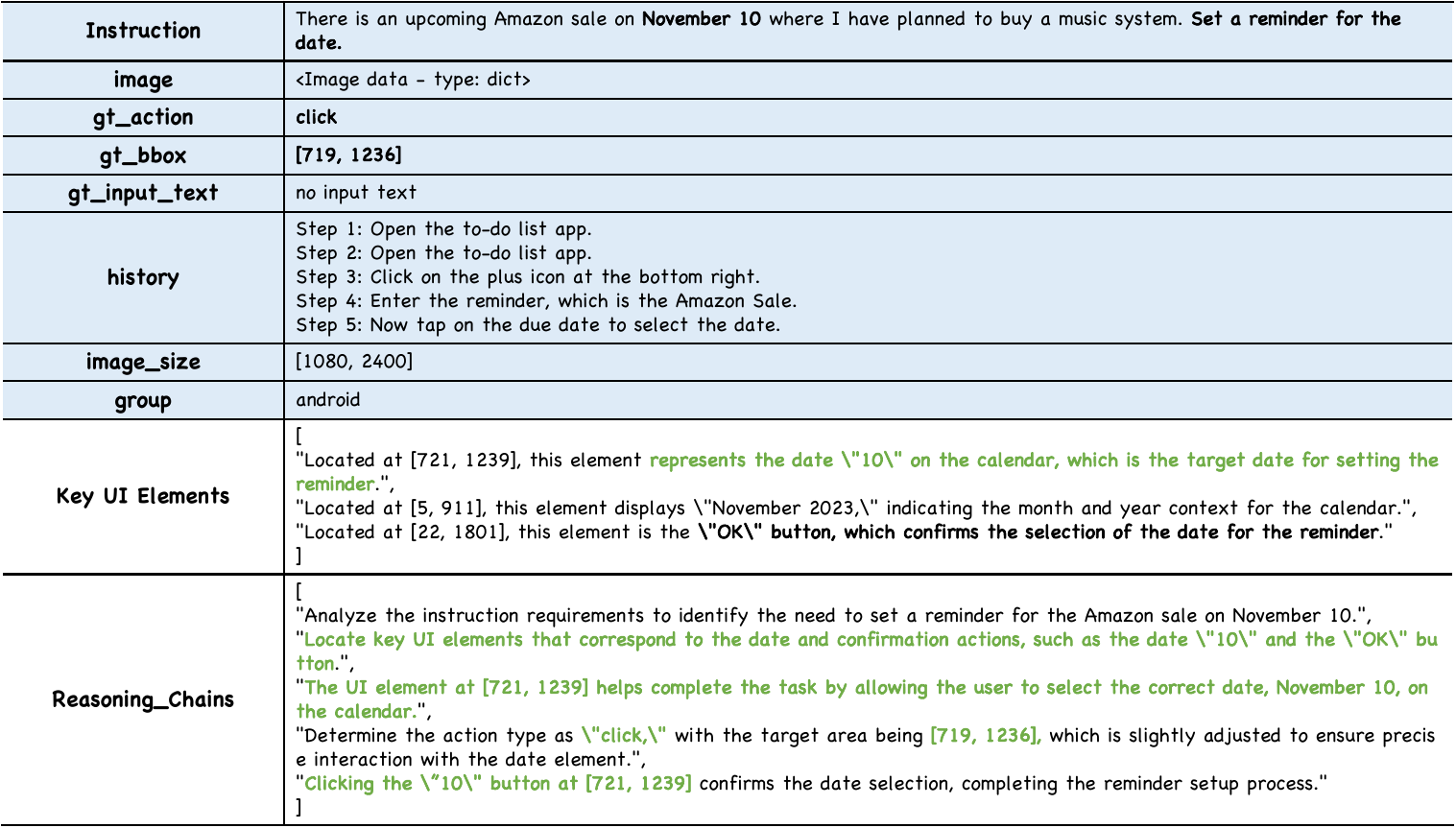} 
    \caption{Case with \textit{click} actions in our UI Comprehension-Bench.}
    \label{fig:case-study3}
\end{figure*}

Meanwhile, we present UI Comprehension-Bench through detailed sample examples. We demonstrate the data fields and values for samples corresponding to common actions including ``open\_app", ``type", and ``click", as shown in Fig. \ref{fig:case-study1}, \ref{fig:case-study2}, \ref{fig:case-study3}. The blue parts indicate the data fields from the existing ``Screen-to-Action" paradigm datasets, whereas our UI Comprehension-Bench additionally incorporates \textit{Key UI Elements} and \textit{Reasoning\_Chains}, which represent the Locate, Lingualize, and Leverage information of UI elements, respectively.

\section{Prompt Details}
\label{Prompt Details}
Since different tasks have different action spaces, we specify the corresponding actions in prompts for each task. We adopt the design principles demonstrated in prior work \citep{li2025layerlogitslogicempowering,Li_2025,liu2026cogcontrollablegraphreasoning} on reasoning prompts.

For GUI grounding tasks (e.g., ScreenSpot-Pro dataset).

\begin{Promptbox}{Prompt for Grounding}
You are UILoop, a reasoning GUI Agent Assistant. In this UI screenshot <image>, I want you to continue executing the command '{text}', with the action history being '{history}'.

Please provide the action to perform (enumerate from ['click']), the point where the cursor is moved to (integer) if a click is performed, and any input text required to complete the action.

Output the location, semantics, and function of UI element(s) that you think are beneficial for reasoning within <ui> </ui> tags, reason based on the key UI element(s) and output the thinking process in <think> </think> tags, and the final answer in <answer> </answer> tags as follows:

<ui> Located at [x, y], describe the UI element’s semantics and function. </ui> <think> ... </think> <answer>[{'action': enum['click'], 'point': [x, y], 'input\_text': 'no input text [default]'}]</answer>

Note: For each UI element, you must provide its location [x, y], semantics, and functionality. Example:

<ui> Located at [743, 724], this element represents the 'Slide Notes' section where users can click to interact with notes related to a slide. </ui>

<ui> Located at [317, 501], this element is a text label that reads \"Developer Tools,\" indicating the section related to developer options. </ui>

Example of answer output:

[{'action': enum['click'], 'point': [123, 300], 'input\_text': 'no input text'}]

\end{Promptbox}

For GUI reasoning tasks (e.g., Android Control-High dataset).

\begin{figure*}[htbp]
    \centering
    \includegraphics[width=1.0\textwidth, keepaspectratio]{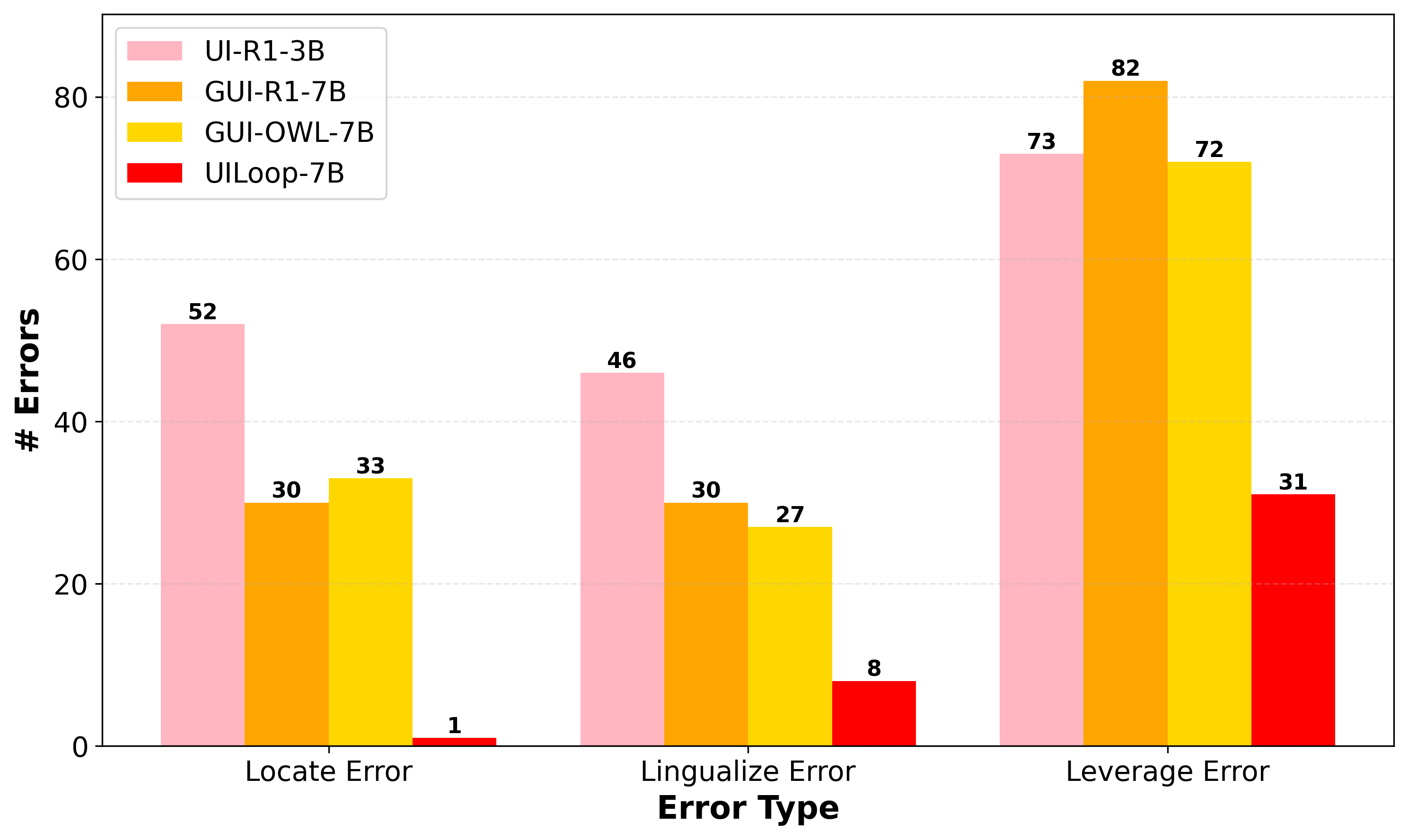} 
    \caption{Error analysis of ``Screen-to-Action" paradigm methods UI-R1-3B, GUI-R1-7B, GUI-OWL-7B and our method UILoop. We demonstrate that the primary error types include: (1) Locate Error, (2) Lingualize Error, and (3) Leverage Error.}
    \label{fig:appendix-error-analysis}
\end{figure*}

\begin{Promptbox}{Prompt for Reasoning}
You are UILoop, a reasoning GUI Agent Assistant. In this UI screenshot <image>, I want you to continue executing the command '{text}', with the action history being '{history}'.

Please provide the action to perform (enumerate from ['wait', 'long\_press', 'click', 'press\_back', 'type', 'open\_app', 'scroll']), the point where the cursor is moved to (integer) if a click is performed, and any input text required to complete the action.

Output the location, semantics, and function of UI element(s) that you think are beneficial for reasoning within <ui> </ui> tags, reason based on the key UI element(s) and output the thinking process in <think> </think> tags, and the final answer in <answer> </answer> tags as follows:

<ui> Located at [x, y], describe the UI element’s semantics and function. </ui> <think> ... </think> <answer>[{'action': enum['wait', 'long\_press', 'click', 'press\_back', 'type', 'open\_app', 'scroll'], 'point': [x, y], 'input\_text': 'no input text [default]'}]</answer>

Note: For each UI element, you must provide its location [x, y], semantics, and functionality. Example:

<ui> Located at [743, 724], this element represents the 'Slide Notes' section where users can click to interact with notes related to a slide. </ui>

<ui> Located at [317, 501], this element is a text label that reads \"Developer Tools,\" indicating the section related to developer options. </ui>

Specific input text (no default) is necessary for actions enum['type', 'open\_app', 'scroll'] Example:

[{'action': enum['wait', 'press\_back'], 'point': [-100, -100], 'input\_text': 'no input text'}]

[{'action': enum['click', 'long\_press'], 'point': [123, 300], 'input\_text': 'no input text'}]

[{'action': enum['type', 'open\_app'], 'point': [-100, -100], 'input\_text': 'shanghai shopping mall'}]

[{'action': enum['scroll'], 'point': [-100, -100], 'input\_text': enum['up', 'left', 'right', 'down']}]

\end{Promptbox}

When employing the selection model (e.g., GPT-4o) to perform Key UI Element Identification and Parsing, we design the prompt as follows.

\begin{Promptbox}{Key UI Element Identification and Parsing}

\# UI Element Analysis and Action Reasoning Task

\#\# Task Description
You need to analyze the given user interface information, identify key UI elements that help complete the specified instruction, and explain how to reason about the correct action based on these elements.

\#\# Input Information
**User Instruction:**
{instruction}

**Action History:**
{history}

**Ground Truth - Action Type:**
{gt\_action}

**Ground Truth - Target Area:**
{gt\_bbox}

**Ground Truth - Input Text:**
{gt\_input\_text}

**UI Element Information:**
{ui\_info}

\#\# Output Requirements

\#\#\# 1. UI Element Functional Descriptions
Please provide a one-sentence description of the UI element's position in the image and its semantic and functional description for each key UI element that helps complete the instruction, with each UI element description enclosed in <ui></ui> tags:

<ui>Located at [x1,y1], this element [semantic and functional description]</ui>
<ui>Located at [x2,y2], this element [semantic and functional description]</ui>
...

\#\#\# 2. Action Reasoning Process
Based on the identified correct UI elements, please explain the reasoning process for deriving the correct action in no more than 5 sentences, with each thought enclosed in <think></think> tags:

<think>
Analyze instruction requirements
</think>

<think>
Locate key UI elements
</think>

<think>
Explain why the UI element(s) help(s) complete the task
</think>

<think>
Determine action type, target area, input text
</think>

<think>
Other necessary thoughts...
</think>

\#\# Important Notes
1. UI element descriptions must be concise and clear, one sentence per element
2. The reasoning process should be logically clear, showing the complete reasoning chain from analysis to decision
3. Strictly follow the specified XML tag format for output
4. Focus on UI elements directly related to completing the instruction

\end{Promptbox}

\section{Error Analysis}

In this section, we conduct a comparative error analysis between current ``Screen-to-Action" paradigm methods UI-R1-3B, GUI-R1-7B, GUI-OWL-7B and our ``Screen-UI Elements-Action" paradigm method. Specifically, we investigate three primary error types related to UI elements: (1) Locate Error, (2) Lingualize Error, and (3) Leverage Error. We randomly sampled 100 instances from the Android Control-High test set and performed manual statistics, as shown in Fig. \ref{fig:appendix-error-analysis}. The results demonstrate that our method achieves error counts of 1, 8, and 31 for Locate, Lingualize, and Leverage Errors respectively, which are substantially lower than the error counts of UI-R1-3B, GUI-R1-7B, and GUI-OWL-7B. This demonstrates the advanced level of our method's mastery over UI elements.

%% file: paper/table/details-benchmark-statistic.tex
\begin{table*}[ht]
\centering
\resizebox{1.0\textwidth}{!}{%
\begin{tabular}{lccccccccc}
\hline
\multicolumn{1}{c}{\multirow{3}{*}{\textbf{Datasets}}} & \multirow{3}{*}{\textbf{\# Episodes}} & \multirow{3}{*}{\textbf{\makecell{\# Unique \\ Instructions}}} & \multicolumn{7}{c}{\textbf{Annotation}} \\ \cline{4-10} 
\multicolumn{1}{c}{} &  &  & \multirow{2}{*}{\textbf{\makecell{Screen \\ Desc.}}} & \multicolumn{3}{c}{\textbf{Key UI Element}} & \multirow{2}{*}{\textbf{\makecell{Action \\ Coord}}} & \multirow{2}{*}{\textbf{\makecell{Action \\ Desc.}}} & \multirow{2}{*}{\textbf{\makecell{Action \\ Think}}} \\
\multicolumn{1}{c}{} &  &  &  & \textbf{Loc.} & \textbf{Lin.} & \textbf{Lev.} &  &  &  \\ \hline
PixelHelp & 187 & 187 & \textcolor{red}{\ding{55}} & \textcolor{red}{\ding{55}} & \textcolor{red}{\ding{55}} & \textcolor{red}{\ding{55}} & \textcolor{green}{\ding{51}} & \textcolor{red}{\ding{55}} & \textcolor{red}{\ding{55}} \\
MoTIF & 4707 & 270 & \textcolor{red}{\ding{55}} & \textcolor{red}{\ding{55}} & \textcolor{red}{\ding{55}} & \textcolor{red}{\ding{55}} & \textcolor{green}{\ding{51}} & \textcolor{red}{\ding{55}} & \textcolor{red}{\ding{55}} \\
UGIF & 523 & 420 & \textcolor{red}{\ding{55}} & \textcolor{red}{\ding{55}} & \textcolor{red}{\ding{55}} & \textcolor{red}{\ding{55}} & \textcolor{green}{\ding{51}} & \textcolor{red}{\ding{55}} & \textcolor{red}{\ding{55}} \\
Meta-GUI & 4684 & 1125 & \textcolor{red}{\ding{55}} & \textcolor{red}{\ding{55}} & \textcolor{red}{\ding{55}} & \textcolor{red}{\ding{55}} & \textcolor{green}{\ding{51}} & \textcolor{green}{\ding{51}} & \textcolor{red}{\ding{55}} \\
AITW & 715142 & 30378 & \textcolor{red}{\ding{55}} & \textcolor{red}{\ding{55}} & \textcolor{red}{\ding{55}} & \textcolor{red}{\ding{55}} & \textcolor{green}{\ding{51}} & \textcolor{red}{\ding{55}} & \textcolor{red}{\ding{55}} \\
GUIAct & 5696 & 5696 & \textcolor{red}{\ding{55}} & \textcolor{red}{\ding{55}} & \textcolor{red}{\ding{55}} & \textcolor{red}{\ding{55}} & \textcolor{green}{\ding{51}} & \textcolor{green}{\ding{51}} & \textcolor{red}{\ding{55}} \\
OmniACT & 9802 & - & \textcolor{red}{\ding{55}} & \textcolor{red}{\ding{55}} & \textcolor{red}{\ding{55}} & \textcolor{red}{\ding{55}} & \textcolor{green}{\ding{51}} & \textcolor{green}{\ding{51}} & \textcolor{red}{\ding{55}} \\
Android Control & 15283 & 15283 & \textcolor{green}{\ding{51}} & \textcolor{red}{\ding{55}} & \textcolor{red}{\ding{55}} & \textcolor{red}{\ding{55}} & \textcolor{green}{\ding{51}} & \textcolor{green}{\ding{51}} & \textcolor{red}{\ding{55}} \\
AITZ & 2504 & 2504 & \textcolor{green}{\ding{51}} & \textcolor{red}{\ding{55}} & \textcolor{red}{\ding{55}} & \textcolor{red}{\ding{55}} & \textcolor{green}{\ding{51}} & \textcolor{green}{\ding{51}} & \textcolor{green}{\ding{51}} \\
MMBench-GUI & 8123 & 8123 & \textcolor{red}{\ding{55}} & \textcolor{green}{\ding{51}} & \textcolor{red}{\ding{55}} & \textcolor{red}{\ding{55}} & \textcolor{green}{\ding{51}} & \textcolor{green}{\ding{51}} & \textcolor{green}{\ding{51}} \\
ScreenSpot & 1272 & 1272 & \textcolor{red}{\ding{55}} & \textcolor{green}{\ding{51}} & \textcolor{red}{\ding{55}} & \textcolor{red}{\ding{55}} & \textcolor{red}{\ding{55}} & \textcolor{red}{\ding{55}} & \textcolor{red}{\ding{55}} \\
ScreenSpot-V2 & 1272 & 1272 & \textcolor{red}{\ding{55}} & \textcolor{green}{\ding{51}} & \textcolor{red}{\ding{55}} & \textcolor{red}{\ding{55}} & \textcolor{red}{\ding{55}} & \textcolor{red}{\ding{55}} & \textcolor{red}{\ding{55}} \\
ScreenSpot-Pro & 1581 & 1581 & \textcolor{red}{\ding{55}} & \textcolor{green}{\ding{51}} & \textcolor{red}{\ding{55}} & \textcolor{red}{\ding{55}} & \textcolor{red}{\ding{55}} & \textcolor{red}{\ding{55}} & \textcolor{red}{\ding{55}} \\
UI-E2I-Bench & 1477 & 1477 & \textcolor{red}{\ding{55}} & \textcolor{green}{\ding{51}} & \textcolor{red}{\ding{55}} & \textcolor{red}{\ding{55}} & \textcolor{red}{\ding{55}} & \textcolor{red}{\ding{55}} & \textcolor{red}{\ding{55}} \\
UI-Vision & 8227 & $\sim$450 & \textcolor{green}{\ding{51}} & \textcolor{green}{\ding{51}} & \textcolor{red}{\ding{55}} & \textcolor{red}{\ding{55}} & \textcolor{green}{\ding{51}} & \textcolor{green}{\ding{51}} & \textcolor{red}{\ding{55}} \\ \hline
\textbf{Ours} & 26207 & 15735 & \textcolor{green}{\ding{52}} & \textcolor{green}{\ding{52}} & \textcolor{green}{\ding{52}} & \textcolor{green}{\ding{52}} & \textcolor{green}{\ding{52}} & \textcolor{green}{\ding{52}} & \textcolor{green}{\ding{52}} \\ \hline
\end{tabular}
}
\caption{Detailed comparison of our UI Comprehension-Bench with existing GUI reasoning benchmarks.}
\label{tab:details-benchmark-statistic}
\end{table*}